\documentclass[10pt,twocolumn,letterpaper]{article}

\usepackage{iccv}
\usepackage{times}
\usepackage{epsfig}
\usepackage{graphicx}
\usepackage{amsmath}
\usepackage{amssymb}

\usepackage[breaklinks=true,bookmarks=false]{hyperref}

\iccvfinalcopy 

\def\iccvPaperID{****} 

\ificcvfinal\fi

\begin{document}
	
	\title{Dynamic Sampling for Deep Metric Learning}
	
	\author{Chang-Hui Liang\\
		Xiamen University\\
		Xiamen, China\\
		{\tt\small chliang@stu.xmu.edu.cn}
		\and
		Wan-Lei Zhao \\
		Xiamen University\\
		Xiamen, China\\
		{\tt\small wlzhao@xmu.edu.cn}
		\and
		Run-Qing Chen\\
		Xiamen University\\
		Xiamen, China\\
		{\tt\small chenrq1010026261@stu.xmu.edu.cn}
	}
	
	\maketitle
	\ificcvfinal\thispagestyle{empty}\fi
	
	\begin{abstract}
		
\footnotetext[1]{\textbf{This paper has been submitted to ``\textit{Pattern Recognition Letters}''}}
Deep metric learning maps visually similar images onto nearby locations and visually dissimilar images apart from each other in an embedding manifold. The learning process is mainly based on the supplied image negative and positive training pairs. In this paper, a dynamic sampling strategy is proposed to organize the training pairs in an easy-to-hard order to feed into the network. It allows the network to learn general boundaries between categories from the easy training pairs at its early stages and finalize the details of the model mainly relying on the hard training samples in the later. Compared to the existing training sample mining approaches, the hard samples are mined with little harm to the learned general model. This dynamic sampling strategy is formularized as two simple terms that are compatible with various loss functions. Consistent performance boost is observed when it is integrated with several popular loss functions on fashion search, fine-grained classification, and person re-identification tasks.
	\end{abstract}

\section{Introduction}
Distance metric learning (usually referred to as metric learning), aims at constructing a task-specific distance measure based on given data. The learned distance metric is then used to support various tasks such as classification, clustering, and retrieval. Conventionally, the metric learning is designed to learn a matrix for the parametric \textit{Mahalanobis} distance. Such that the similar contents are close to each other under the learned \textit{Mahalanobis} distance, while the distance between the dissimilar contents is large.

Due to the great success of deep learning in many computer vision tasks in recent years, it has been gradually introduced to metric learning, which is widely known as deep metric learning. Instead of learning the distance metric directly, deep metric learning learns feature embedding from the raw data. For instance, given images $x_a$, $x_b$ and $x_c$, $x_a$ and $x_b$ are from the same category while $x_c$ is distinct from them. The deep metric learning learns a non-linear mapping function $\mathcal{F}(\cdot)$ that embeds $x_a$, $x_b$ and $x_c$ to the new feature space. In this embedding space, $\mathcal{F}(x_a)$ and $\mathcal{F}(x_b)$ are close to each other, and $\mathcal{F}(x_c)$ is dissimilar to both of them under a predefined distance metric $\mathit{m(\cdot,\cdot)}$.

\begin{figure*}[t]
	\begin{center}
		\includegraphics[width=0.8\linewidth]{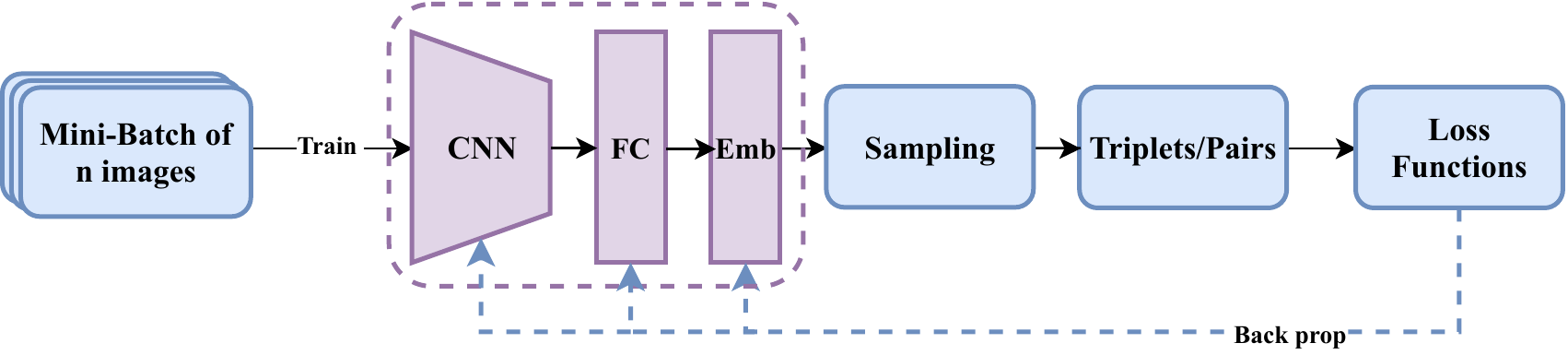}
	\end{center}
	\caption{The general framework of deep metric learning. The training images are organized into mini-batches. Images are then forward to a pre-trained ConvNets. The \textit{d}-dimensional feature of one is produced by the `Emb' layer, which is an extra fully-connected layer attached to FC-layer. The distances between image pairs are aggregated into a single loss value by a pre-defined loss function. The embedding space is optimized by iteratively minimizing this function loss.}
	\label{fig:framework}
\end{figure*}

Owing to the seminal learning framework from~\cite{chopra2005learning}, deep metric learning has been successfully adopted in various tasks such as online fashion search~\cite{huang2015cross,liu2016deepfashion,hadi2015buy,ji2017cross}, face recognition~\cite{chopra2005learning,schroff2015facenet}, person re-identification~\cite{yi2014deep}, and fine-grained image search~\cite{sohn2016improved,oh2016deep,wang2019multi}, etc. In general, the embedding space is learned on image pairs/triplets driven by loss functions. Namely, the training images are organized into positive pairs (images from the same category) and negative pairs (images from different categories). The loss function is designed to distill all the pair-based category information into a single loss value. The training process aims to build an embedding space by minimizing this loss. Such that the pairwise relations reconstructed in the embedding space coincide well with that of being  supplied to the training. Since the number of pairwise relations is quadratic to the size of training image set, it is computationally expensive to enumerate all the pairwise relations of the training set. As a consequence, the definition of loss function along with the ushered-in pair-sampling strategy becomes critical. The general framework of deep metric learning is shown in Fig.~\ref{fig:framework}.

In the literature, a series of loss functions have been proposed one after another.  \textit{Contrastive loss}~\cite{hadsell2006dimensionality} and \textit{triplet loss}~\cite{triplet} are the two most popular loss functions. However, both of them fail to make full use of the pairwise relations in a mini-batch. In addition, it is widely observed that the large portion of image pairs are easy training samples. Hard training samples, which take up a small portion, are more decisive to the category boundaries. Due to the lack of strategy to mine on these hard training samples, deep metric learning based solely on  \textit{contrastive loss} and \textit{triplet loss} converges slowly. To alleviate this issue, \textit{N-Pair loss}~\cite{sohn2016improved}, \textit{lifted structure loss}~\cite{oh2016deep}, and \textit{multiple similarity loss}~\cite{wang2019multi} consider more pairwise relationships within one mini-batch. This leads to a much faster convergence pace and better discriminativeness of the learned embedding space. The performance is further boosted by the mining of the hard negatives in~\cite{sohn2016improved,oh2016deep,wang2019multi,ge2018deep,roth2019mic}.

In this paper, a simple but effective dynamic training sample mining strategy is proposed to boost the performance of deep metric learning. The idea is inspired by two observations, which largely diverge from the common believes in deep metric learning. On the one hand, it is commonly agreed that the images from the same category should be close to each other. However, we observe that it is  harmful to the model if they are pushed as close as possible. The model could suffer from overfitting when the hard positives are pushed too close to each other. Hard negatives face the similar problem. On the other hand, it is true that the hard training samples are more informative than the easy ones. In order to attain a more discriminative model, these hard samples should be fed to the training with high priority. However, it will be more effective when they are fed to the training at the later stage, while leaving the process to learn from relatively easy examples at its early stage. 

Based on the first observation, we only pull/push the positive/negative pairs above/below a similarity threshold to alleviate the over-fit issue. Based on the second observation, the impact of hard training samples is tuned dynamically, which is planted in the design of a loss function. Namely, unlike current practices in \textit{N-Pair loss}~\cite{sohn2016improved}, \textit{lifted structure loss}~\cite{oh2016deep}, or \textit{multiple similarity loss}~\cite{wang2019multi}, the hard training samples take higher effect as the training epoch grows. This mimics the cognitive process of human beings that learns general concepts from simple cases and drills deeper into the complex cases step by step. Considerable improvement is observed as this dynamic sampling strategy is integrated with \textit{lifted structure loss}, \textit{multiple similarity loss}, \textit{triplet loss}, as well as \textit{binomial deviance loss}. According to our experiments, such kind of improvement is consistent across different tasks such as fashion search, person re-identification, and fine-grained image search. 

The remainder of this paper is organized as follows. Section~\ref{sec:relat} reviews the most representative loss functions in deep metric learning. Our dynamic sampling strategy is presented in Section~\ref{sec:mthd}. The comparative study over the representative loss functions and the proposed sampling strategy is presented in Section~\ref{sec:exp}. Section~\ref{sec:conc} concludes the paper.


\section{Related Work}
\label{sec:relat}
In this section, several representative loss functions and the enhancement strategies over them in deep metric learning are reviewed. In order to facilitate our later discussions, several concepts are defined. Given a pair of images $\{x_i, x_j\}$, the distance between them is given as
\begin{equation}
s_{i, j} = m(\mathcal{F}(x_i), \mathcal{F}(x_j)),
\end{equation}
where $m(\cdot,\cdot)$ is a pre-defined distance measure. It could be \textit{Cosine} similarity or \textit{Euclidean} distance, etc. For clarity, the following discussion is made based on \textit{Cosine} similarity by default. Correspondingly, the label for this image pair is given as $y_{i,j}$. Positive pair is given as $y_{i,j}=1$, which indicates $x_i$ and $x_j$ come from the same category. While $y_{i,j}$ equals to $0$ when they come from different categories. In the deep metric learning literature, the loss functions are mostly defined based on the positive and negative pairs, which are organized into a series of mini-batches for the sake of training efficiency.

\textit{Contrastive loss}~\cite{hadsell2006dimensionality} encodes the similarities from both positive pairs and negative pairs in one loss function. Basically, it regularizes the similarities between positive pairs to be larger than the similarities from negative pairs with a constant margin $\lambda$. Namely,
\begin{equation}
\mathcal{L}_c=\frac{1}{m}\sum_{(i,j)}^{m/2}\Bigg(-y_{i,j} s_{i, j}+(1-y_{i,j})[0, s_{i, j} - \lambda]_+\Bigg),
\label{eqn:lc}
\end{equation}
where $m$ is the number of anchor-positive pairs in one training batch. $[\cdot]_+$ in Eqn.~\ref{eqn:lc} is the hinge loss. In order to guarantee that there are sufficient anchor-positive pairs in one batch, a fixed number of positive pairs are selected for one batch. The loss function neglects the fact that the dissimilar scales of two images to another image could be different. The minimization on $\mathcal{L}_c$ tends to converge as long as the distance between one pair satisfies with the margin $\lambda$\footnote{$\lambda$ acts as the minimum margin between positives and negatives for all the loss funtions discussed in the paper.}.  The loose constraint over the pairwise distance leads to slow convergence.

\textit{Binomial deviance loss} (\textit{BD-loss})~\cite{yi2014deep} can be viewed as a soft version of \textit{contrastive loss}. Its loss function is given as
\begin{equation}
\begin{aligned}
\mathcal{L}_{b} = \sum_{i=1}^{m} \Bigg( &\frac{1}{P} \sum_{y_{a,b}=1} \log{\left[ 1+\rm e^{\alpha(\lambda - s_{a,b})} \right]}+\\ &\frac{1}{N} \sum_{y_{c,d} =0} \log{\left[ 1+\rm e^{\beta(s_{c,d} - \lambda)} \right]}\Bigg),
\end{aligned}
\label{eqn:lb}
\end{equation}
where $P$ and $N$ are the numbers of positive pairs and negative pairs respectively. $\alpha$ and $\beta$ in Eqn.~\ref{eqn:lb} are the scaling factors. Compared to hinge loss function in Eqn.~\ref{eqn:lc}, softplus function in Eqn.~\ref{eqn:lb} demonstrates similar shape as hinge loss while is smooth and with continuous gradients. According to recent studies~\cite{yi2014deep,manandhar2019semantic}, it shows superior performance in person re-identification, fashion search and fine-grained image search tasks.

In order to enhance the discriminativeness of the embedding space, \textit{triplet loss} (defined in Eqn.~\ref{eqn:lt}) is designed to maximize the similarity between an anchor-positive pair $\{x_{a},x_{p}\}$ in contrast to the similarity from anchor to a negative sample $x_{n}$. 
\begin{equation}
\mathcal{L}_t=\frac{3}{2m}\sum_{i=1}^{m/2}[s_{a,n}^{(i)}-s_{a,p}^{(i)}+\lambda]_+
\label{eqn:lt}
\end{equation}
Compared to \textit{contrastive loss}, the gap between positive and negative pairs is defined in terms of relative similarity. Namely, the relative similarity between the positive and negative pair with respect to anchor. It therefore remains effective for various scenarios compared to that of \textit{contrastive loss}. Nevertheless, it is expensive to enumerate all possible triplets in one mini-batch. As a result, some informative training samples may be left unused if no particular mining schemes are introduced. 

In the literature, several efforts~\cite{schroff2015facenet,oh2016deep,ge2018deep,roth2019mic} have been made to mine on the hard training samples to further boost its performance. In order to make full use of the samples inside one mini-batch, \textit{lifted structure loss} (see Eqn.~\ref{eqn:lf})~\cite{oh2016deep} is proposed. The loss function is designed to consider all the positive and  negative pairs in one mini-batch.
\begin{equation}
\mathcal{L}_{f} = \sum_{i=1}^{m} \bigg[\log{\sum_{y_{i,j}=1}\rm e^{\lambda - s_{i,j}}}  + \log{\sum_{y_{i, j}=0} \rm e^{s_{i,j}}} \bigg]_+
\label{eqn:lf}
\end{equation}
The objective of \textit{lifted structure loss} is to maximize the margin between negatives and positives. The negatives which are the closest to positive images are mined during the training.  Because the optimization on nested maximum function converges to bad local optimum, the loss function is relaxed to optimize a smooth one. 


Essentially, there are three types of negative pairs in the pair-based learning. Namely, they are the anchor-negative pairs, positive-negative pairs and negative-negative pairs. In the aforementioned models, they usually consider one or two of them while missing another. \textit{Multiple similarity loss}~\cite{wang2019multi} is proposed to consider the similarities from all three types of negative pairs. The loss function is given in Eqn.~\ref{eqn:lm}.
\begin{equation}
\begin{aligned}
\mathcal{L}_{m} =\frac{1}{m}\sum_{i=1}^{m} \Bigg( &\frac{1}{\alpha} \log{[ 1+\sum_{k \in \mathcal{P}_i}\rm e^{-\alpha(s_{i,k} - \lambda)} ]}+\\ &\frac{1}{\beta} \log{[ 1+\sum_{k \in \mathcal{N}_i}\rm e^{\beta(s_{i,k} - \lambda)} ]}\Bigg)
\end{aligned}
\label{eqn:lm}
\end{equation}
where $\alpha$ and $\beta$ are the scaling parameters. In the above loss function, $\mathcal{P}_i$ and $\mathcal{N}_i$ are the positive and negative image sets respectively. Similar as \textit{BD-loss}, the loss function is defined based on softplus function.  In this learning model, the hard negatives are mined from the second type of pairs. The selected training samples are then weighted by the similarities from the other two types of pairs. According to~\cite{wang2019multi}, similar performance as \textit{BD-loss} is reported.

In the literature,  there are many efforts have been taken to enhance the performance of aforementioned models. For instance, online sampling is one of the typical ways. Conventionally, training samples are paired in advance, which might lead to over-fitting. In~\cite{sohn2016improved,oh2016deep,wang2019multi,wang2019ranked}, the training images are organized into pairs/triplets during the training so that one anchor can be paired with more number of samples. 

Besides the loss function and sampling strategies, the batch size and scaling parameters in the loss function (\textit{e.g.}, $\alpha$ in Eqn.~\ref{eqn:lb}) are the other two factors impact the performance. Larger batch size allows the loss function  to consider more pairwise/triangular relationships among images. This in turn leads to a better embedding space structure. However, large batch size also leads to higher computation complexity. As a result, a trade-off has to be made. The scaling factors are the few hyper-parameters have to be tuned manually in many loss functions such as \textit{BD-loss} and \textit{multiple similarity loss}. According to recent study~\cite{wang2019multi}, the parameter tuning alone could lead to \textit{20\%} performance improvement.

In our solution, the training samples selection is novelly abstracted to weighting terms and integrated with the loss function. It allows the training process to learn the embedding space in an order, namely from relatively easy samples to the harder. Therefore, the hard concepts (carried by hard samples) are learned without overwriting the learned general (easy) concept. Moreover, this scheme is generic in the sense it could be integrated with various loss functions. Its effectiveness is confirmed on fashion search, person re-identification, and fine-grained image search when it is integrated with \textit{lifted structure loss}, \textit{multiple similarity loss}, \textit{triplet loss}, and \textit{BD-loss}.

\section{Dynamic Training Pair Selection}
\label{sec:mthd}
In this section, our strategies that are designed to boost the performance of deep metric learning are presented. In general, most of the deep metric learning approaches are defined based on training image pairs. The variations across different approaches mainly lie in the selection of training pairs and the definition of loss function. In this section, we first present the heuristics we used in the training pair selection. Based on the heuristics, various existing loss functions, that are integrated with two dynamic sampling terms, are presented.
\begin{figure*}[t]
	\begin{center}
		\includegraphics[width=0.6\linewidth]{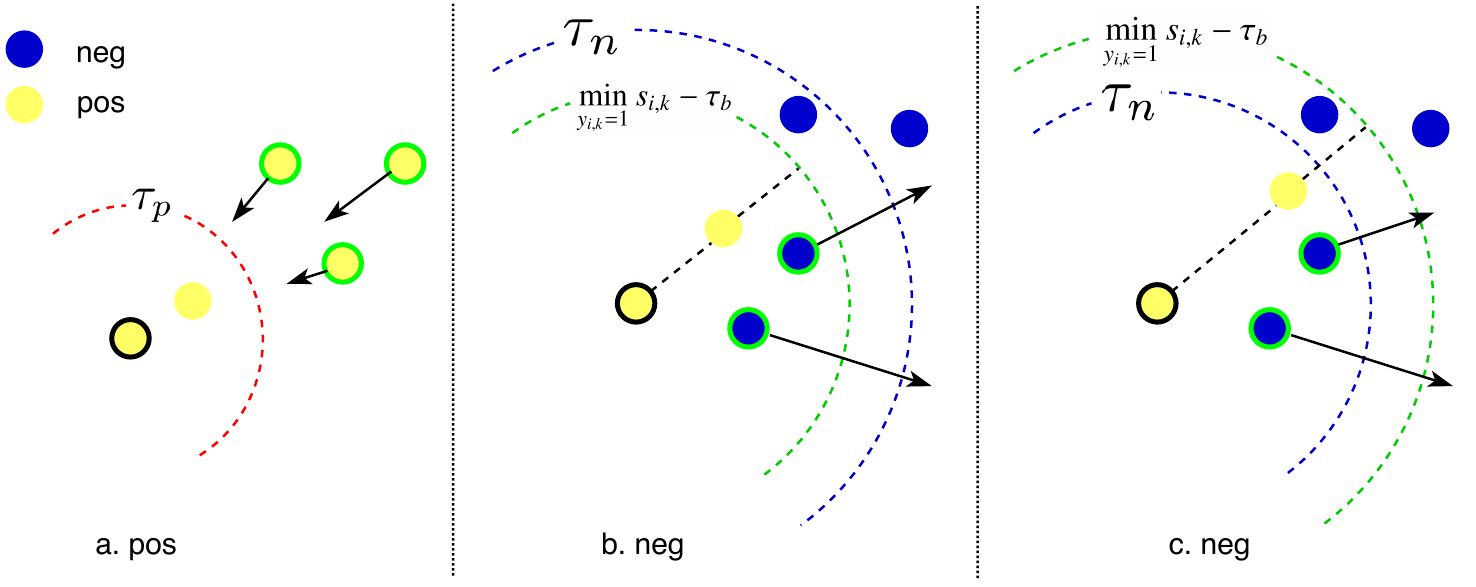}
	\end{center}
	\caption{The illustration of the role of three thresholds. The dashed circles in red, blue, and green represent borders regularized by $\tau_p$, $\tau_n$ and Inequation~\ref{eqn:tb} respectively. In figure (a), the positive samples which meet with $s_{i,k} < \tau_p$ will be selected. For negative pairs, they could be in either cases illustrated in figure (b) or figure (c). Under figure (b) and (c) cases, the negative pair $\{i,j\}$ that $s_{i,j} > \tau_n$ and $s_{i,j} > \min_{y_{i,k}} s_{i,k} - \tau_b$ will be selected.}
	\label{fig:threshold}
\end{figure*}

\subsection{Heuristics in Pairs Mining}
As witnessed in many research works~\cite{oh2016deep,wang2019multi,ge2018deep,roth2019mic}, the easy training samples take a large portion in a mini-batch, however they are less helpful  than the hard training pairs. Similar as other works~\cite{wang2019ranked}, hard thresholds are set to filter out these easy training samples. To achieve that, two similarity thresholds $\tau_p$ and $\tau_n$, namely one for easy positives and another for easy negative pairs, are introduced. Given the similarity between a positive pair $\{x_i, x_j\}$ is $s_{i,j}$, positive pair $\{x_i, x_j\}$ will not be considered in the training if  $s_{i,j} \geqslant \tau_p$. Similarly, a negative pair $\{x_k, x_m\}$ is not considered in the training as  $s_{k,m} \leqslant \tau_n$. In our implementation, $\tau_p$ and $\tau_n$ are set to \textit{0.9} and \textit{0.1} respectively. As the training continues, the boundary between positives and negatives becomes clearer. One could imagine more and more negative and positive pairs will be filtered out by these two thresholds and therefore will no longer join in the training.

In addition to $\tau_p$ and $\tau_n$, similar as~\cite{wang2019multi}, a flexible margin is set for negative pairs to separate them from the most remote positive pairs. Namely, given a positive pair $\{x_i, x_k\}$ and a negative pair $\{x_i, x_j\}$, $\{x_i, x_j\}$ will be selected to join in the training when the following inequation holds.

\begin{equation}
s_{i,j} > \min_{y_{i,k}=1} s_{i,k} - \tau_b,
\label{eqn:tb}
\end{equation}
where $\tau_b$ is the lower bound similarity of a positive sample to the anchor.

The roles that these three thresholds take are illustrated in Fig.~\ref{fig:threshold}. On the one hand, threshold $\tau_p$ prevents the training from pushing the positives as close as possible. On the other hand, threshold $\tau_n$ prevents the negative pairs from being pulled too far away. These two thresholds together prevent  the structure of the learned embedding space from collapsing due to overfitting. According to our observation, very few training samples could pass through these two thresholds at the early training stage. As the training continues for several rounds, more and more negative pairs and positives are well separated in the embedding space. They are, therefore, set aside by these two thresholds. The training gets focus more and more on harder training samples. The flexible threshold given by Inequation~\ref{eqn:tb} takes similar effect. As one could imagine, $\min_{y_{i,k}=1} s_{i,k}$ is relatively small at the early training stage. As the embedding space evolves to a better structure,  $\min_{y_{i,k}=1} s_{i,k}$ grows bigger. This in turn thresholds out more and more relatively easy training samples. 

\subsection{Dynamic Metric Learning Loss}
Based on the above heuristics, only relatively hard pairs are joined in the training. Among these relatively hard training samples, the degree of hardness still varies from one training pair to another. In the ideal scenario, it is expected that the training samples are organized sequentially according to the degree of hardness. Samples with low degree of hardness are fed to the training at the early stages. As the training model evolves, harder training samples are fed to the training process since they become more critical to define the category borders. Intuitively, one has to prepare a group of image pairs for training with increasing degree of hardness each time. Although it sounds plausible, it is hard to operate as the hardness degree of one image pair varies along with the evolving embedding space. In the following, a novel weighting strategy based on image pair similarity $s_{i,j}$ is proposed. It regularizes the importance of a training pair according to its hardness in the training. The easy training samples are assigned with higher importance at the early training stages and the importance of hard ones grows as the training epoch increases.

In the existing loss functions, the similarity between one pair of image $s_{a,b}$ is mainly designed to weight how much penalty we should be aggregated into the loss function. In our design, it is additionally used to indicate the degree of a pair joined in one round of training. Specifically, for a positive pair $\{a, b\}$, the value $\tau_p - s_{a,b}$ basically indicates the hardness degree. The larger this value is, the harder the positive pair is. The value $s_{c,d} - \tau_n$ has the similar efficacy for a negative pair $\{c,d\}$. Let's take \textit{BD-loss} as an example. We show how the training samples are re-weighted according to their degree of hardness. Given $E_t$ is the number of total epochs we need to train our model, the current number of epochs that the training has been undertaken is given as $E_c$ ($ 1 \leq E_c \leq E_t$). Two terms $\frac{2E_c}{E_t}(\tau_p - s_{a,b})^2$ and $\frac{2E_c}{E_t}(s_{c,d} - \tau_n)^2$, one for positive and one for negative, are introduced to \textit{BD-loss}. The loss function in Eqn.~\ref{eqn:lb} is rewritten as 
\begin{equation}
\begin{aligned}
\mathcal{L}^*_b = \sum_{i=1}^{m} \Bigg\{ &\frac{1}{P} \sum_{y_{a,b}=1} \log{\left[ 1+\rm e^{\alpha[(\lambda - s_{a,b}) + \frac{2E_c}{E_t}(\tau_p - s_{a,b})^2 ]} \right]}+\\ &\frac{1}{N} \sum_{y_{c, d}=0} \log{\left[ 1+\rm e^{\beta[(s_{c,d} - \lambda) + \frac{2E_c}{E_t}(s_{c,d} - \tau_n)^2 ]}  \right]}\Bigg\}.
\end{aligned}
\label{eqn:lb+}
\end{equation}
Apparently, these two terms are impacted by both ${E_c}$ and the image pair similarity $s$. On the one hand, the larger the gap between $s$ and the corresponding bound (either $\tau_p$ or $\tau_n$) is, the higher these two terms are. This basically indicates the degree of hardness for a training pair (either positive or negative). Hard pairs tend to hold high weights. On the other hand, the terms are also controlled by the number of current epochs. The more number of epochs the training is undertaken, the higher of impact these two terms have on the overall loss $\mathcal{L}^*_b$. This leads the optimization to focusing on these hard training pairs more and more as $E_c$ grows bigger. 

Similarly, for \textit{lifted structure loss} (Eqn.~\ref{eqn:lf}), \textit{triplet loss} (Eqn.~\ref{eqn:lt}) and \textit{multiple similarity loss} (Eqn.~\ref{eqn:lm}), they are rewritten as Eqn.~\ref{eqn:lf+}, Eqn.~\ref{eqn:lt+} and Eqn.~\ref{eqn:lm+} if these two terms are integrated.
\begin{equation}
\begin{aligned}
\mathcal{L}^*_f = \sum_{i=1}^{m} \Bigg\{&\log{\sum_{y_{a,b}=1}\rm e^{[(\lambda - s_{a,b}) + \frac{2E_c}{E_t}(\tau_p - s_{a,b})^2 ]}} \\ &+ \log{\sum_{y_{c,d}=0} \rm e^{[s_{c,d}+ \frac{2E_c}{E_t}(s_{c,d} - \tau_n)^2 ]}} \Bigg\}_+
\end{aligned}
\label{eqn:lf+}
\end{equation}

\begin{equation}
\begin{aligned}
\mathcal{L}^*_t=\frac{3}{2m}\sum_{i=1}^{m/2}\Bigg\{&\frac{1}{P}\sum_{y_{a,b}=1}[- s_{a,b} + \frac{2E_c}{E_t}(\tau_p - s_{a,b})^2] + \\&\frac{1}{N}\sum_{y_{c,d}=0}[s_{c,d} + \frac{2E_c}{E_t}(s_{c,d} - \tau_n)^2] + \lambda
\Bigg\}_+
\end{aligned}
\label{eqn:lt+}
\end{equation}

\begin{equation}
\begin{aligned}
\mathcal{L}^*_m =\frac{1}{m}\sum_{i=1}^{m} \Bigg\{ &\frac{1}{\alpha} \log{[ 1+\sum_{a \in \mathcal{P}_i}\rm e^{[-\alpha(s_{i,a} - \lambda) + \frac{2E_c}{E_t}(\tau_p - s_{i,a})^2]} ]}+\\ &\frac{1}{\beta} \log{[ 1+\sum_{b \in \mathcal{N}_i}\rm e^{[\beta(s_{i,b} - \lambda) + \frac{2E_c}{E_t}(s_{i,c} - \tau_n)^2]} ]}\Bigg\}
\end{aligned}
\label{eqn:lm+}
\end{equation}

\begin{figure}[t]
	\begin{center}
		\includegraphics[width=0.75\linewidth]{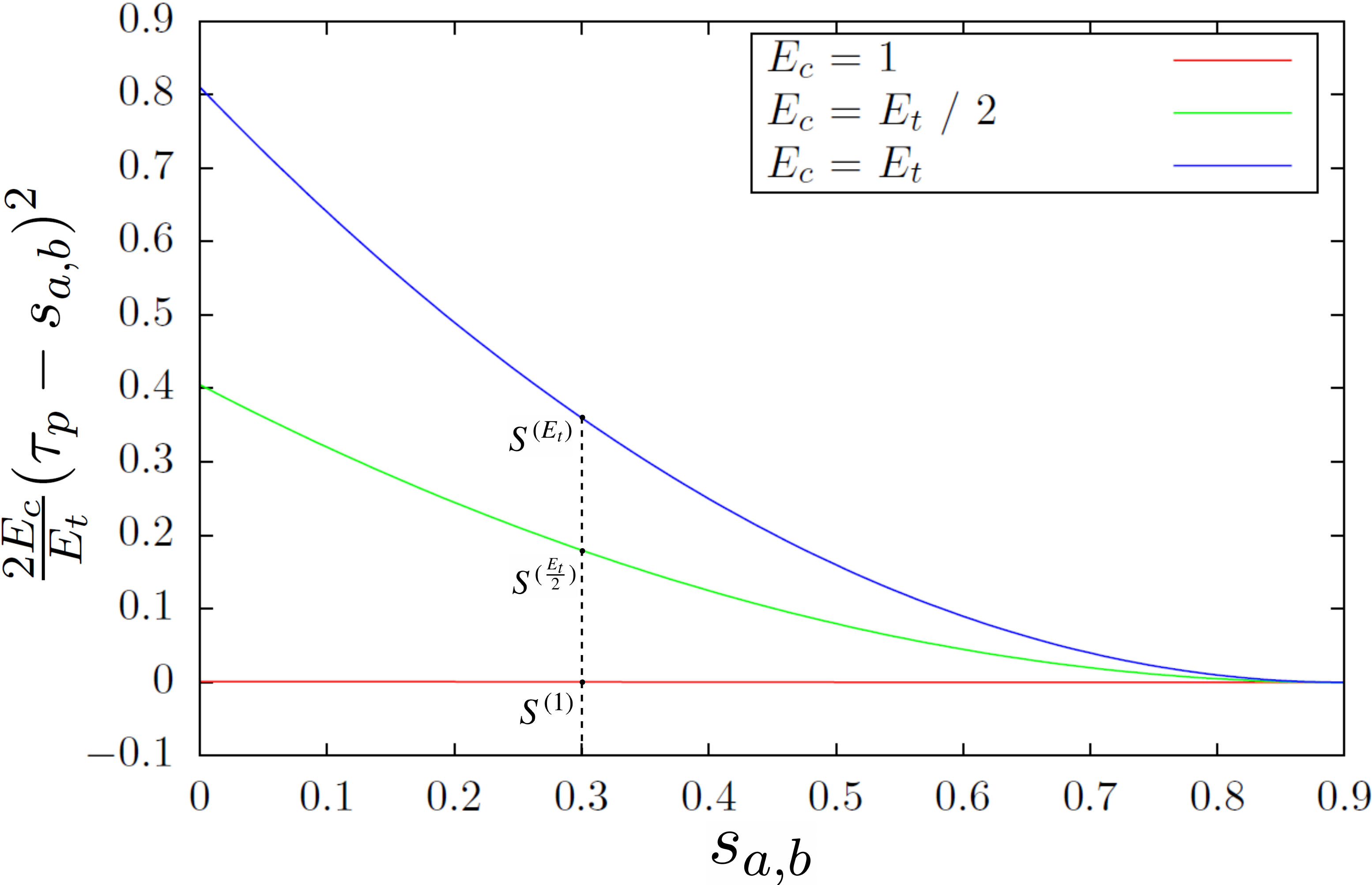}
	\end{center}
	\caption{The illustration of re-weighting term for positive pairs. The term is given as the function of similarity $s_{a,b}$. This term assigns different weights to different positive pairs according to their mutual similarities. The weights vary as the epoch $E_c$ grows. The figure shows the re-weighting curves when training is at \textit{1}st, $E_t/2$-th and $E_t$-th epoch. As $E_c$ grows, re-weighting is biased towards hard training pairs (whose $s_{a,b}$ is low). This leads to the higher contribution to the final loss.}
	\label{fig:term}
\end{figure}

In Eqn.~\ref{eqn:lf+}, Eqn.~\ref{eqn:lt+},  and Eqn.~\ref{eqn:lm+}, terms $\frac{2E_c}{E_t}(\tau_p - s_{a,b})^2$ and $\frac{2E_c}{E_t}(s_{c,d} - \tau_n)^2$ play a similar role as they do on \textit{BD-loss}. Fig.~\ref{fig:term} shows the weights produced by term $\frac{2E_c}{E_t}(\tau_p - s_{a,b})^2$ for positive pair with respect to $s_{a,b}$. The similarity of a positive pair $s_{a,b}$ is in the range of $[0.0, 0.9]$ after thresholding by $\tau_p$. Low $s_{a,b}$ indicates the positive pair is close to the category boundary, namely it is a hard positive pair. As shown in the figure, the weights we assign to all the pairs are equally low at the first epoch of the training. The impact of this term on the loss function is therefore minor. Since the easy pairs take a large portion in one mini-batch, the training is actually biased towards easy pairs at the early stages. As epoch $E_c$ grows, higher weights are assigned to pairs with lower similarities (as shown by the green curve in Fig.~\ref{fig:term}). The bias towards hard positives is more significant as epoch $E_c$ grows even bigger, which in turn leads to the higher contribution from hard positives to the loss function. Similar thing happens to the term for negative pairs. As a consequence, the optimization on the above models is tuned to focusing on the hard pairs gradually as the epoch grows.


As will be revealed in the experiment section, the simple modification on these popular loss functions leads to considerable performance enhancement.  The performance wins over the original model could be as high as more than \textit{100\%}.

The advantages of such modification are three folds. Firstly, the training samples are fed to the model from easy to hard as the training epoch grows. This allows the training process to focus on relatively easy samples at the early stage and hard samples at its later stage.  Secondly, this training sample selection rule is integrated with the loss function, no extra complexity is induced. Moreover, it is a generic strategy as it is compatible with different types of loss functions.

\subsection{Implementation Details}
The pre-trained Inception~\cite{ioffe2015batch} is employed as the backbone network for our deep metric learning. Each training image is first fed into the network to generate a fixed-length feature vector, which maps the image into the embedding space. The size of the feature vector is typically set to \textit{512} in our study.  Thereafter, the images are organized into mini-batches. In each training epoch, \textit{25} classes are randomly selected. Five images are randomly selected from each of these classes. This results in \textit{125} images in one mini-batch. Thereafter, the pairwise similarity within this mini-batch is calculated. The image pairs which pass through the thresholds $\tau_p$, $\tau_n$ and satisfy with Inequation~\ref{eqn:tb} are selected to join in the training. To this end, the loss is computed based on the revised loss function. For instance, Eqn.~\ref{eqn:lb+} is employed for \textit{BD-loss}. The computed loss for one mini-batch is back-propagated to optimize the network.  $\alpha$, $\lambda$, and $\beta$ in Eqn.~\ref{eqn:lb+} are set to  \textit{2}, \textit{0.5}, and \textit{40} respectively. The above training process loops for $E_t$ rounds.

For \textit{lifted structure loss}, \textit{triplet loss}, and \textit{multiple similarity loss}, the training process remains largely the same. The only difference lies in the loss function. For \textit{lifted structure loss}, \textit{triplet loss}, and \textit{multiple similarity loss}, Eqn.~\ref{eqn:lf+}, Eqn.~\ref{eqn:lt+} and Eqn.~\ref{eqn:lm+} are employed respectively.  $\lambda$ in Eqn.~\ref{eqn:lf+} is set to \textit{1.0}. $\lambda$ in Eqn.~\ref{eqn:lt+} is set to \textit{0.5}. $\alpha$, $\lambda$, and $\beta$ in Eqn.~\ref{eqn:lm+} are set to  \textit{2}, \textit{0.5}, and \textit{50} respectively. All the codes are implemented with PyTorch and are available on GitHub\footnote{https://github.com/CH-Liang/DSDML}.

\begin{table*}[t]
	\footnotesize{
		\caption{Summary over five datasets used in the evaluation}
		\label{tab:data}
		\begin{center}
			\begin{tabular}{|l|llll|l|}
				\hline
				Datasets & \#Images & \#Train & \#Query & \#Test/Validation  & Task \\
				\hline\hline
				In-shop~\cite{liu2016deepfashion} & 52,712 & 25,882 & 14,218 & 12,612 & Fashion search\\
				\hline
				Deepfashion2~\cite{ge2019deepfashion2} & 224,389 & 191,961 & 10,990 & 21,438 & Fashion search\\
				\hline
				CUB200~\cite{wah2011caltech}   & 11,788  & 5,864 & - & 5,924 & fine-grained search \\
				\hline
				Cars-196~\cite{krause20133d}   & 16,185  & 8,054 & - & 8,131 & fine-grained search \\
				\hline
				Market-1501~\cite{zheng2015scalable}   & 36,036  & 12,936 & 3,368 & 19,732 & Re-Identification \\
				\hline
			\end{tabular}
	\end{center}}
\end{table*}

\section{Experiments}
\label{sec:exp}
In this section, the effectiveness of the proposed dynamic hard training sample mining strategy is studied when it is integrated with four popular loss functions. Namely, they are \textit{BD-loss} (BD),  \textit{triplet loss} (TP), \textit{multiple similarity loss} (MS), and \textit{lifted structure loss} (LF). They are denoted as BD*, TP*, MS*, and LF* respectively  when the proposed dynamic sampling strategy is integrated. The study is made on three different tasks, namely fashion search, fine-grained image search, and person re-identification. The behavior of these popular loss functions on these three tasks are comprehensively studied. The performance from each loss function is also compared to state-of-the-art approaches in each particular task.

\subsection{Datasets and Evaluation Protocols}
For fashion search, datasets In-Shop (also known as Deepfashion)~\cite{liu2016deepfashion} and Deepfashion2~\cite{ge2019deepfashion2} are adopted in the evaluation.
Dataset \textit{In-Shop} is a collection of fashion product images crawled from online shopping websites. There are \textit{52,712} images covering across \textit{7,982} classes. \textit{14,218} images are set aside as the queries. For dataset \textit{Deepfashion2}, it is comprised of images both from online shops and users. Considerable portion of the images are directly collected from \textit{Deepfashion}. 
For \textit{Deepfashion2}, the retrieval is defined as user-to-shop query. Namely, the images uploaded by the users are treated as queries. The same product images crawled from online shopping websites are treated as search targets. It is more challenging than  \textit{In-Shop} task as it is a cross-domain search problem. Since the test set for \textit{Deepfashion2} is not released yet, the validation set is treated as the candidate dataset for search evaluation. 

For fine-grained image search, \textit{CUB-200-2011}~\cite{wah2011caltech} and \textit{CARS196}~\cite{krause20133d} are adopted, both of which are the most popular evaluation benchmarks in deep metric learning. \textit{CUB-200-2011} covers \textit{200} bird categories. Following the conventional practice,  \textit{5,864} images of \textit{100} categories are used for training, and the remaining \textit{5,924} images from \textit{100} categories are used for evaluation. Dataset \textit{CARS196}, is comprised of \textit{16,185} car images that cover across \textit{196} categories. The \textit{8,054} images from \textit{98} classes are used for training. The rest \textit{8,131} images from the same group of classes are used for testing.

On person re-identification task, the experiments are conducted on \textit{Market-1501}~\cite{zheng2015scalable} dataset. It consists of \textit{32,668} images from \textit{1,501} individuals. They are captured by \textit{6} cameras of different viewpoints. In the training, \textit{12,936} images from \textit{750} individuals are used. This setting is fixed for all the evaluation approaches. The brief information about all the five datasets are summarized in Tab.~\ref{tab:data}.

In our implementation, images from all datasets are resized to $256{\times}256$ and then randomly cropped to $224 {\times} 224$. For data augmentation, we followed the configurations in ~\cite{oh2016deep}. Specifically, random cropping and random horizontal flips are employed during training and single cropping is employed during testing. The backbone network for all the four loss functions we considered is Inception~\cite{ioffe2015batch}. The Adam optimizer is adopted in the training. Following the convention in the literature, we report our performance in terms of Recall@K. To be line with the evaluation convention on different benchmarks, difference series of Ks are taken on different datasets. All the experiments are pulled out on a server with NVIDIA GTX \textit{1080} Ti GPU setup.

\begin{table}
	\footnotesize{
		\begin{center}
			\caption{Ablation analysis of BD-loss on Deepfashion2 validation set}
			\label{tab:dp2ab}
			\begin{tabular}{|l|c|cccc|}
				\hline
				Recall@ & Dim. & 1 & 5 & 10 & 20 \\
				\hline\hline
				BD & 512 & 40.2 & 55.4 & 62.7 & 70.1  \\
				BD+T & 512 & 41.6 & 57.0 & 63.7 & 70.5  \\
				BD+W & 512 & 42.9 & 58.5 & 65.7 & 72.8\\
				$BD^*$ & 512  &\textbf{44.0} & \textbf{60.2} & \textbf{67.2} & \textbf{73.7} \\
				
				\hline
			\end{tabular}
		\end{center}
	}
\end{table}

\subsection{Ablation Study with Fashion Search}
Before we show the performance improvement that the overall dynamic sampling strategy brings for different loss functions, the study on the contribution of each step in dynamic sampling to the final improvement  is presented. The study is typically conducted with \textit{BD-loss} on the fashion search task. However, the similar trend is observed on other loss functions and other tasks. In the experiment, the \textit{BD-loss} that is integrated with simple thresholding with $\tau_p$, $\tau_n$ and Inequation~\ref{eqn:tb} is given as ``BD+T''. While \textit{BD-loss} integrated with two re-weighting terms only is given as ``BD+W''. \textit{BD-loss} integrated with both is given as $BD^*$. The results of these three runs on \textit{Deepfashion2} are shown in Tab.~\ref{tab:dp2ab}. The result from \textit{BD-loss} is treated as the comparison baseline.

As shown in the table, both schemes achieve consistent improvement. On average, the dynamic sampling brings more considerable improvement than simple thresholding. Specifically, more than \textit{2\%} improvement is observed on `BD+W'' run across different $K$s. The best performance is observed when two schemes are integrated as a whole. This basically indicates they are complementary to each other. In the following experiments, all the four loss functions we study here are integrated with these two enhancement schemes.


\subsection{Fashion Search}
In this section, the effectiveness of the proposed enhancement strategy on four loss functions is studied in fashion search task. Representative approaches in the literature on this task are considered in the study. FashionNet~\cite{liu2016deepfashion} and Match R-CNN~\cite{ge2019deepfashion2} are treated as the comparison baselines for \textit{In-Shop} and \textit{Deepfashion2} respectively. They are proposed along with these two benchmarks. The fashion search is treated as a sub-task under the multi-task learning framework. In addition, recent deep metric learning approaches Divide and Conquer (Divide)~\cite{sanakoyeu2019divide}, Mining Interclass Characteristics (MIC)~\cite{roth2019mic}, \textit{Angular loss} (Angular)~\cite{wang2017deep}, \textit{Batch hard triplet loss} (Hard Triplet)~\cite{hermans2017defense} and \textit{N-Pair loss} (N-Pair)~\cite{sohn2016improved} are also considered in the comparison. Among these approaches, MIC learns auxiliary encoder for the visual attributes, which induces extra computational costs. Attention-based Ensemble (ABE)~\cite{kim2018attention} is the representative approach of ensemble deep metric learning. 
%

The performance on \textit{In-Shop} and \textit{Deepfashion2} is shown on Tab.~\ref{tab:inshop} and Tab.~\ref{tab:deep2} respectively. As shown in Tab.~\ref{tab:inshop} and Tab.~\ref{tab:deep2}, all the four loss functions which are integrated with dynamic sampling terms demonstrate considerable performance improvement. The improvement ranges from \textit{10-100\%} for different loss functions. Enhanced by the dynamic sampling, the performance from all loss functions become competitive to or even outperforms the most effective approach in the literature, in particular on the challenging dataset \textit{Deepfashion2}. An interesting observation is that the performance difference between different loss functions becomes much smaller when being all supported by the dynamic sampling. Overall, \textit{BD-loss} with dynamic sampling shows the best performance on the two datasets. The performance from BD$^*$ is similar as the one reported in~\cite{manandhar2019semantic}, which however requires product attributes in the training.


\begin{table}[h]
	\footnotesize{
		\begin{center}
			\caption{Comparison with the state-of-the-art approaches on \textit{In-Shop}}
			\label{tab:inshop}
			\begin{tabular}{|l|r|cccccc|}
				\hline
				Recall@ & Dim. & 1 & 10 & 20 & 30 & 40 & 50\\
				\hline\hline
				FashionNet$^{\ddag}$~\cite{liu2016deepfashion} & 4,096& 53.0 & 73.0 & 76.0 & 77.0 & 79.0 & 80.0 \\
				Divide$^{\ddag}$~\cite{sanakoyeu2019divide}&128 & 85.7 & 95.5 & 96.9 & 97.5 & - & 98.0\\
				ABE$^{\ddag}$~\cite{kim2018attention}&512 & 87.3 & 96.7 & 97.9 & 98.2 & 98.5 & 98.7\\
				MIC$^{\ddag}$~\cite{roth2019mic}&128 & 88.2 & 97.0 & - & 98.0 & - & 98.8\\
				\hline
				BD&512 & 87.3 & 96.1 & 97.4 & 97.9 & 98.2 & 98.4 \\
				BD$^*$& 512 & \textbf{89.8} & \textbf{97.6} & \textbf{98.3} & \textbf{98.6} & \textbf{98.8} & \textbf{98.9}\\
				\hline
				LF&512        & 35.3 & 65.5 & 73.6 & 77.9 & 80.6 & 82.5\\
				LF$^*$ & 512 & 81.5 & 93.8 & 95.7 & 96.5 & 97.1 & 97.4\\
				\hline
				MS & 512        & 87.3 & 96.3 & 97.3 & 98.0 & 98.3 & 98.5\\
				MS$^*$&512 & 87.7 & 96.7 & 97.6 & 98.1 & 98.4 & 98.7 \\	
				\hline
				TP & 512    & 83.7 & 94.5 & 96.2 & 97.0 & 97.4 & 97.7\\
				TP$^*$&512 & 84.7 & 95.1 & 96.6 & 97.2 & 97.6 & 97.9 \\
				\hline
			\end{tabular}\\
			$^{\ddag}$ digits are cited from the referred paper.
		\end{center}
	}
	
\end{table}

\begin{table}[h]
	\footnotesize{
		\caption{Comparison with the state-of-the-art approaches on \textit{Deepfashion2} validation set}
		\label{tab:deep2}
		\begin{center}
			\begin{tabular*}{\hsize}{@{}@{\extracolsep{\fill}}|l|c|cccc|@{}}
				\hline
				Recall@ &Dim. & 1 & 5 & 10 & 20 \\
				\hline\hline
				Match R-CNN$^{\ddag}$~\cite{ge2019deepfashion2}&256 & 26.8 & - & 57.4 & 66.5  \\
				Angular$^{\ddag}$~\cite{wang2017deep}&128 & 32.4 & 47.9 & 55.3 & 62.3\\
				Hard Triplet$^{\ddag}$~\cite{hermans2017defense}&128 & 32.4 & 48.9 & 56.0 & 63.2  \\
				N-Pair$^{\ddag}$~\cite{sohn2016improved}&128 & 32.8 & 50.1 & 57.9 & 64.8 \\
				MIC~\cite{roth2019mic}&512 & 38.1 & 52.1 & 59.3 & 66.3 \\
				Divide~\cite{sanakoyeu2019divide}&512 & 39.4 & 54.4 & 61.5 & 68.5 \\
				\hline
				BD&512 &  40.2 & 55.4 & 62.7 & 70.1  \\
				BD$^*$&512 &\textbf{44.0} & \textbf{60.2} & \textbf{67.2} & \textbf{73.7} \\
				\hline
				LF&512 & 15.7 & 28.3 & 35.3 & 43.3  \\
				LF$^*$&512 & 39.3 & 55.5 & 63.3 & 70.6 \\
				\hline
				MS&512 & 41.0 & 55.9 & 62.7 & 70.2  \\
				MS$^*$&512 & 42.4  & 58.0  & 65.2  & 72.5 \\
				\hline
				TP&512 & 38.6 & 54.1 & 61.6 & 68.9  \\
				TP$^*$&512 & 41.2 & 57.0  &  64.4 & 71.9 \\
				\hline
			\end{tabular*}\\
			$^{\ddag}$ digits are cited from the referred paper.
		\end{center}
	}
	
\end{table}

\begin{figure*}
	\begin{center}
		\includegraphics[width=0.85\linewidth]{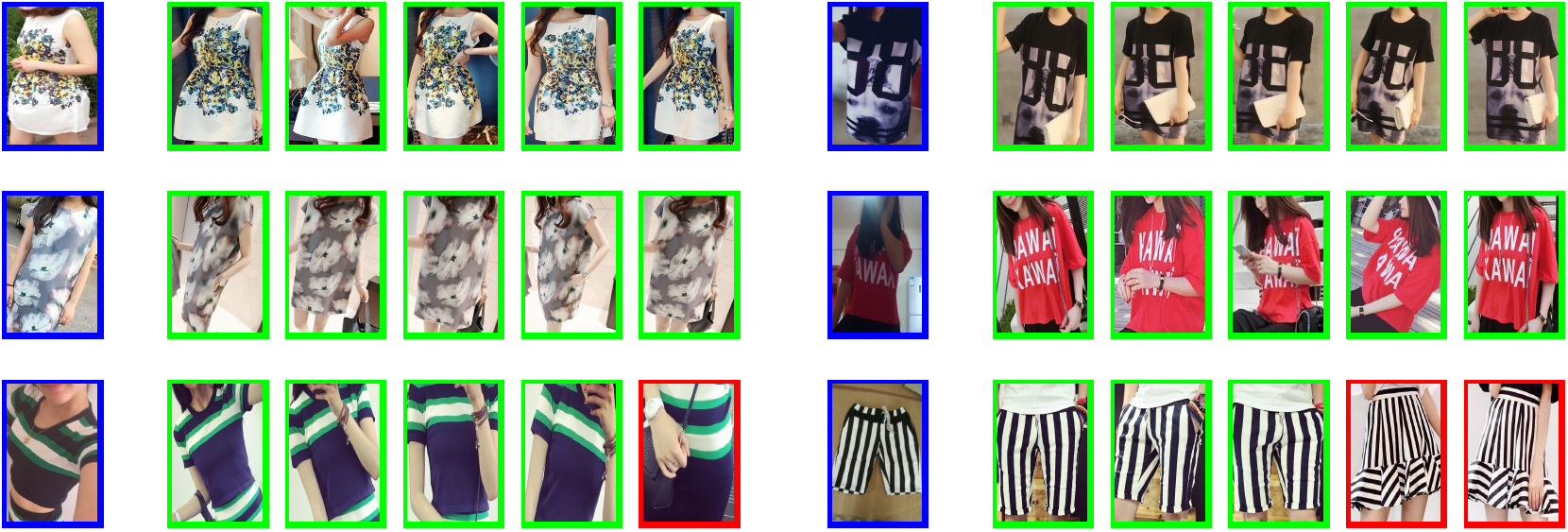}
	\end{center}
	\caption{Top-5 search results on \textit{Deepfashion2}. The queries are shown on the first column and followed by retrieved top-5 candidates on each row.}
	\label{fig:fashrslt}
\end{figure*}

The search result samples from BD$^*$ on \textit{Deepfashion2} are shown in Fig.~\ref{fig:fashrslt}. As shown in the figure, there are some false positives, nevertheless they are reasonable in the sense that they look very similar to the queries.  Moreover, our approach shows steady performance even under severe variations in illumination and image quality.

\subsection{Fine-grained search}
In this section, the performance of the enhanced loss functions is evaluated on \textit{CUB-200-2011} and \textit{CARS196} for fine-grained search. Three state-of-the-art deep metric learning approaches Divide~\cite{sanakoyeu2019divide}, ABE~\cite{kim2018attention} and MIC~\cite{roth2019mic} are considered in the comparison. The performance results on these two datasets are shown in Tab.~\ref{tab:cub} and Tab.~\ref{tab:cars}. respectively.


The performance from all enhanced loss functions show consistent improvement over the original ones. The improvement is particularly significant on \textit{lifted structure loss}. BD$^*$ and MS$^*$ perform competitively well with the state-of-the-art approaches, namely Divide and MIC. Compared to Divide and MIC, our approach is much lightweight. Approach Divide requires to learn several sub-embedding spaces. Similarly, extra computation is required in MIC to train the auxiliary encoder for the latent visual attributes. In contrast, considerable improvement from our approach is achieved by injecting two re-weighting terms into the loss function. No sophisticated modification on the training process or the network architecture is required.



\begin{table}[h]
	\footnotesize{
		\begin{center}
			\caption{Comparison with the state-of-the-art approaches on \textit{CUB200}}
			\label{tab:cub}
			\begin{tabular}{|l|c|cccccc|}
				\hline
				Recall@&Dim. & 1 & 2 & 4 & 8 & 16 & 32\\
				\hline\hline
				ABE$^{\ddag}$~\cite{kim2018attention}&512& 60.6 & 71.5 & 79.8 & 87.4 & - & -\\
				Divide$^{\ddag}$~\cite{sanakoyeu2019divide}&128& 65.9 & 76.6 & 84.4 & \textbf{90.6} & - & -\\
				MIC$^{\ddag}$~\cite{roth2019mic}&128& \textbf{66.1} & \textbf{76.8} & \textbf{85.6} & - & - & -\\
				\hline
				BD & 512 & 63.5 & 74.5 & 82.8 & 88.9 & 93.6 & 96.6 \\
				BD$^*$&512 & 63.8 & 75.1 & 83.3 & 89.7 & \textbf{94.1} & 96.8\\
				\hline
				LF & 512        & 47.2 & 59.1 & 70.3 & 80.3 & 87.2 & 92.4\\
				LF$^*$ & 512 & 54.9 & 66.7 & 77.3 & 84.8 & 90.6 & 94.6\\
				\hline
				MS & 512        & 64.3 & 75.0 & 83.3 & 89.6 & 93.9 & 96.8\\
				MS$^*$& 512 & 64.8 & 75.1 & 84.0 & 90.3 & \textbf{94.1} &\textbf{96.9} \\	
				\hline
				TP & 512        & 52.4 & 63.9 & 73.7 & 82.4 & 89.4 & 93.6\\
				TP$^*$ & 512 & 55.2 & 66.3  & 76.5 &84.5  &  90.5& 94.4 \\	
				\hline
			\end{tabular}\\
			$^{\ddag}$ digits are cited from the referred paper.
		\end{center}
	}
\end{table}

\begin{table}[h]
	\caption{Comparison with the state-of-the-art approaches on \textit{Cars-19}6}
	\footnotesize{
		\begin{center}
			\begin{tabular}{|l|c|cccccc|}
				\hline
				Recall@&Dim. & 1 & 2 & 4 & 8 & 16 & 32\\
				\hline\hline
				ABE$^{\ddag}$~\cite{kim2018attention}&512&\textbf{85.2}  & 90.5 & 94.0 & 96.1 & - & -\\
				Divide$^{\ddag}$~\cite{sanakoyeu2019divide}&128& 84.6 &\textbf{90.7} &\textbf{94.1}  & \textbf{96.5} & - & -\\
				MIC$^{\ddag}$~\cite{roth2019mic}&128& 82.6 & 89.1 & 93.2 & - & - & -\\
				\hline
				BD & 512 & 79.7 & 86.9 & 91.9 & 95.1 & 97.1 & 98.4 \\
				BD$^*$& 512 & 81.2 & 88.3 & 92.8 & 95.8 & \textbf{97.6} &\textbf{98.8}\\
				\hline
				LF & 512        & 37.9 & 49.8 & 61.4 & 72.0 & 81.7 & 88.6\\
				LF$^*$ & 512 & 63.6 & 73.5 & 81.4 & 87.8 & 92.4 & 95.5\\
				\hline
				MS & 512  & 81.1  & 88.0 & 92.6 & 95.4 & 97.4 & 98.6\\
				MS$^*$ & 512 & 81.6 & 88.3 & 92.9 & 95.8 & \textbf{97.6} & \textbf{98.8}\\	
				\hline
				TP & 512        & 55.4  & 66.5 & 75.6 & 83.4 & 89.1 & 93.4\\
				TP$^*$ & 512 & 63.2 & 73.5 & 81.6 & 87.7 & 92.4 & 95.4\\	
				\hline
			\end{tabular}\\
			$^{\ddag}$ digits are cited from the referred paper.
		\end{center}
	}
	\label{tab:cars}
\end{table}

\subsection{Person Re-Identification}
In this section, we further study the effectiveness of the proposed strategy on person re-identification task. The experiment is conducted on \textit{Market-1501}. State-of-the-art approaches BoW+CN~\cite{zheng2015scalable}, Pose-driven Deep Convolutional  (PDC)~\cite{su2017pose} and Part-based Convolutional Baseline  (PCB)~\cite{sun2018beyond} are considered in the comparison. BoW+CN is proposed along with dataset \textit{Market-1501}. It is the color names descriptor (CN) quantized with bag-of-word (BoW). BoW+CN is treated as comparison baseline in this study. Both PDC and PCB build feature representations on body-parts level. Namely, sub-feature vector is learned on each body part (\textit{e.g.}, head and legs), which requires fine-grained annotations as well. The performance from all the approaches is shown in Tab.~\ref{tab:market}.





On the one hand, it is clear to see the consistent improvement of the proposed strategy brings to various loss functions. In particular, the improvement is significant for \textit{lifted structure loss}. All the approaches with the enhanced loss functions outperform the comparison baseline. On the other hand, there is a large performance gap between these deep metric learning based models and the state-of-the-art approaches of person re-identification. This is not surprising given more complex model and more training information are involved for these two state-of-the-art approaches. In contrast, only category information is capitalized in deep metric learning approaches. Nevertheless, this experiment does confirm that the improvement achieved by dynamic sampling  is consistent across various visual retrieval tasks. It also shows how well deep metric learning approach alone could achieve on this complicated task.


\begin{table}[h]
	\footnotesize{
		\begin{center}
			\caption{Comparison with the state-of-the-art approaches on Market-1501 for person re-identification task}
			\label{tab:market}
			\begin{tabular}{|l|r|cccc|}
				\hline
				Recall@&Dim. & 1 & 5 & 10 & 20 \\
				\hline\hline
				BoW+CN$^{\ddag}$~\cite{zheng2015scalable}  &100 &47.3& - &-  &-\\
				PDC$^{\ddag}$~\cite{su2017pose} &2,048 &84.1& 92.7 &94.9  &\textbf{96.8} \\
				PCB$^{\ddag}$~\cite{sun2018beyond} &12,288 &\textbf{93.8}& \textbf{97.5} & \textbf{98.5} & - \\
				\hline
				BD&512 & 60.9 & 82.6& 88.2 & 92.1  \\
				BD*&512 & 61.0& 83.3 & 89.4 & 92.3 \\
				\hline
				LF&512        & 31.0& 53.2 & 62.3 & 71.3 \\
				LF*&512 & 52.0 & 74.5 & 81.7 & 87.8 \\
				\hline
				MS &512       & 62.0& 83.7 & 89.3 & 92.6  \\
				MS*&512 & 63.0 & 84.4 & 89.6 & 93.2   \\	
				\hline
				TP &512       & 49.4& 73.0 & 79.8 & 86.3  \\
				TP*&512 & 53.5& 76.4 &83.1  & 87.8  \\	
				\hline
			\end{tabular}\\
			$^{\ddag}$ digits are cited from the referred paper.
		\end{center}
	}	
\end{table}

Overall, \textit{BD-loss} and \textit{multiple similarity loss} show much more superior performance over other loss functions on three different tasks. With the proposed dynamic sampling strategy, both models demonstrate performance boosts on three different tasks and across different parameter settings. In particular, the performance from the enhanced \textit{BD-loss} and \textit{multiple similarity loss} is better than or close to state-of-the-art approaches on fashion search and fine-grained search tasks.


\section{Conclusion}
\label{sec:conc}

We have presented a simple but effective dynamic sampling strategy to boost the performance of deep metric learning. In our solution, the dynamic sampling is formularized as two terms that are compatible with various loss functions. These two re-weighting terms dynamically tune the impact that a training pair contribute to the loss function. Higher weights are assigned to the easy training pairs at the early training stages. Hard pairs are assigned with increasingly higher weights as the training epoch grows. This allows the network to learn the concepts from easy to hard, which is comparable to the cognitive process of human beings. Experiments on five datasets of three different tasks show consistent performance improvement when this strategy is integrated with four popular loss functions. In addition, we find that \textit{BD-loss} and \textit{multiple similarity loss} are consistently better on three different tasks than the other loss functions by large performance margins.

\bibliographystyle{ieee_fullname}
\bibliography{chliang.bib}
	
\end{document}